\documentclass[11pt,journal, one column]{IEEEtran}
\usepackage{graphicx}
\usepackage{amssymb} 
\usepackage{epstopdf}
\usepackage[ruled,norelsize]{algorithm2e}
\makeatletter
\newcommand{\removelatexerror}{\let\@latex@error\@gobble}
\makeatother
\usepackage[caption=false,font=footnotesize]{subfig}
\usepackage{pseudocode}
\usepackage{fancybox}
\usepackage{multicol}
\usepackage{verbatim}
\usepackage{amsthm}
\usepackage{amsmath}
\usepackage{color}
\usepackage{bm}
\usepackage{cite}
\usepackage{multirow}
\usepackage{tikz}
\usepackage{hyperref}
\usepackage{booktabs}
\usepackage{marginnote}
\newcommand{\mbf}[1]{{\mathbf #1}}
\newcommand{\mbb}{\mathbb}

\begin{document}
%
\title{Bare Demo of IEEEtran.cls\\ for IEEE Journals}

\title{Recovery of damped exponentials using structured low rank matrix completion}
\author{Arvind Balachandrasekaran,~\IEEEmembership{Student Member,~IEEE,} Vincent Magnotta and Mathews Jacob,~\IEEEmembership{Senior Member,~IEEE}
\thanks{Arvind Balachandrasekaran, Mathews Jacob are with the Department of Electrical and Computer Engineering and Vincent Magnotta is with the Department of Radiology, University of Iowa, Iowa City, IA, 52245 USA (e-mail:arvind-balachandrasekaran@uiowa.edu;vincent-magnotta@uiowa.edu;mathews-jacob@uiowa.edu)}
\thanks{This work is supported by grants NIH 1R01EB019961-01A1 and ONR N00014-13-1-0202.}}
\maketitle

\begin{abstract}
We introduce a structured low rank matrix completion algorithm to recover a series of images from their under-sampled measurements, where the  signal along the parameter dimension at every pixel is described by a linear combination of exponentials. We exploit the exponential behavior of the signal at every pixel, along with the spatial smoothness of the exponential parameters to derive an annihilation relation in the Fourier domain. This relation translates to a low-rank property on a structured matrix constructed from the Fourier samples. We enforce the low rank property of the structured matrix as a regularization prior to recover the images. Since the direct use of current low rank matrix recovery schemes to this problem is associated with high computational complexity and memory demand, we adopt an iterative re-weighted least squares (IRLS) algorithm, which facilitates the exploitation of the convolutional structure of the matrix. Novel  approximations involving two dimensional Fast Fourier Transforms (FFT) are introduced to drastically reduce the memory demand and computational complexity, which facilitates the extension of structured low rank methods to large scale three dimensional problems. We demonstrate our algorithm in the MR parameter mapping setting and show improvement over the state-of-the-art methods. 
\end{abstract}


\IEEEpeerreviewmaketitle

\section{Introduction}
Recovering parameters of a linear combination of damped exponentials is a problem of high significance in many MR imaging applications, including MR parameter mapping \cite{ugander2012myocardial, borthakur2008t1}, MR spectroscopy \cite{di2007proton}, and fat/water imaging \cite{hardy1995separation}. The objective is to estimate from a series of MR images the spatial maps of the exponential parameters, which are indicative of the underlying tissue microstructure or metabolism. These maps are often used as bio-markers for pathologies including neuro-degenerative and cardiovascular disorders \cite{ugander2012myocardial, borthakur2008t1, di2007proton}. Current approaches involve acquiring multiple images by sampling the exponential signals at different points, followed by a pixel-by-pixel fitting of the exponential model to estimate the parameters. However, the main challenge with these schemes is the long acquisition time, resulting from the need to acquire a large number of high spatial resolution images. Recently, several researchers have considered compressive sensing methods for recovering images from under-sampled Fourier measurements using priors which enforce sparsity, smoothness and low-rankness \cite{feng2011accelerating, velikina2013accelerating, petzschner2011accelerating, huang2012t2, zhao2015accelerated, zhao2014model, doneva2010compressed, zhang2015accelerating, zhu2015panda, bhave2015accelerated, li2012fast}.  

The recovery of exponential parameters from few uniform samples of a linear combination of exponentials is a classical and well researched problem in signal processing \cite{stoica_spec}. The model has been extended to include a large class of signals with finite rate of innovation or finite number of discontinuities in \cite{FRI-Vett,liang1989high}. The early work in this direction focussed on the uniform sampling setting, where the linear dependencies between the samples of the signal translated to an ``annihilation relationship''. This implies that the signal can be nulled by the convolution with a finite impulse response filter. Recently, several researchers have extended the framework to recover a linear combination of undamped sinusoids from a few non-uniform Fourier samples \cite{chen2014robust, loraks, gregsiam}. These methods compactly represent the annihilation relation as a product of a Hankel matrix formed from the signal samples and a vector of annihilation filter coefficients. With this reformulation, the Hankel matrix can be shown to be low-rank; the low-rank property can be enforced to complete the matrix from its non-uniform measurements.  


In this paper, we introduce a structured matrix completion algorithm for recovering a series of MR images from their non-uniformly under-sampled Fourier measurements, where the signal along the parameter dimension at every pixel is described by a linear combination of damped exponentials. We also assume that the exponential parameters vary smoothly in space. We observe that this model is general enough to account for many applications, including MR spectroscopy, parameter mapping, and diffusion MRI. For example, in MR parameter mapping, the signal along the parameter dimension could vary as a function of echo time, repetition time, and/or spin-lock duration. In this paper, we consider the single parameter setting for simplicity. We exploit the exponential behavior at every pixel, along with the smoothness of the parameters in the spatial dimensions, to derive an annihilation relation in the  $k-t$ domain; $t$ denotes the parameter dimension. These 3-D convolution relations can be compactly represented using a multi-fold Toeplitz matrix formed from the  $k-t$ samples. We show that this matrix has a large null space, and hence is low rank. We enforce the low rank property of the structured matrix as a prior to recover the missing entries from the under-sampled Fourier measurements. The spatial smoothness as well as the number of exponentials in the model can be controlled by the rank of the Toeplitz matrix, which is in-turn dependent on the regularization parameter.

The straightforward implementation of the above structured Toeplitz matrix recovery scheme is associated with huge memory demand and high algorithmic complexity. Specifically, the size of the Toeplitz matrix is often several orders of magnitude greater than the size of the multidimensional signal. We introduce an algorithm based on a half-circulant approximation of the Toeplitz matrix, which eliminates the need for the explicit evaluation and storage of the structured matrix. This work is a generalization of our recent work \cite{GIRAF}, where we introduced the GIRAF (Generalized Iterative Reweighted Annihilating Filter) algorithm for recovering the missing entries of a Toeplitz/Hankel matrix, when only few of its entries are observed. The approximation of linear convolutions by circular convolutions enabled an efficient implementation of the algorithm using Fast Fourier transforms (FFTs). The circulant approximation in \cite{GIRAF} is valid when the signal samples decay rapidly towards the boundaries. This scheme is not directly applicable in our setting, since the signal samples have significant magnitude at the first few points along the parameter dimension. We modify the GIRAF algorithm and adopt a hybrid approach to solve the problem. Specifically, we perform the 3-D linear convolution as a series of 2-D circular convolutions along the spatial dimensions and linear convolution along the parameter dimension. Such a modification allows us to apply our algorithm on large scale multi-dimensional exponential estimation problems. The preliminary version of this work is accepted as a conference paper \cite{AB_ISBI2017}. Compared to the work \cite{AB_ISBI2017}, the theoretical and algorithmic frameworks are further developed here, in addition to the application of the problem to the recovery of single and multi-channel $T_2$ weighted images.

The proposed method has similarities with structured matrix priors introduced to exploit various signal properties, including finite support and smoothly varying phase \cite{loraks}, piece-wise smooth continuous domain images \cite{gregsiam}, and continuous domain wavelet sparsity \cite{ALOHA}. Similar structured low rank priors have also appeared in the recovery of calibrationless multichannel data \cite{shin2014calibrationless} and multi-shot diffusion weighted images \cite{mani2016multi}. However, none of the above 2-D methods are designed to exploit the smooth exponential structure of the 3-D dataset. In \cite{ALOHAPM}, a Hankel matrix is constructed by exploiting the temporal smoothness using a Fourier transform and the spatial redundancies are exploited using a wavelet transform; the exponential structure of the temporal signal is not taken into account. In addition, the recovery of each $k_{y}-t$ slice is performed independently, assuming Cartesian sampling. The exponential structure of the signal is exploited in \cite{MORASA} and \cite{LORA}, where a Hankel matrix is constructed at every pixel by exploiting the linear predictability of the exponential time series. Since the linear combination of pixel-wise structured low rank priors is not capable of exploiting the similarities between the pixels in the dataset, the authors additionally use low rank and joint sparsity penalties on the Casorati matrix; see section II.D for more details. The proposed formulation enables the joint exploitation of the spatial correlations as well as the exponential structure, thereby mitigating the need for additional spatial priors; this approach is computationally more efficient and requires fewer free regularization parameters.

\subsection{Notation}
We collect the different notations used through out the paper and describe them in this section for easy reference. Unless otherwise mentioned, bold upper-case letters $\mathbf X$ and bold lower-case letters $\mathbf y$ are used to represent matrices  and vectors respectively; $\boldsymbol\left[\mathbf{y}\right]^T,$ $\boldsymbol\left[\mathbf{X}\right]^T$ represent a transpose of the vector $\mathbf{y}$ and matrix $\mathbf{X}$ respectively. We denote a function that is dependent on $\mathbf{r}$ and $n$ by $x[\mathbf{r},n]$. The collection of function values for all possible values of $\mathbf r$ and $n$ are denoted by the vector $\mathbf x$. The discrete Fourier transform of $x$ is denoted by $\hat x$, while the vector corresponding to the function values is denoted by $\hat{\mathbf x}$. We use non-bold lower-case greek alphabets such as $\mu, \alpha$ to represent constants. Upper-case greek alphabets $\Lambda, \Theta$ represent index sets containing the support of the coefficients of the filter and Fourier data; $|\Lambda|$ is used to denote the size of the set $\Lambda$. We denote the 2D and 3D convolution by $*$ and $\otimes$ respectively. The calligraphic letters (e.g. $\mathcal A, \mathcal T$) denote operators. For example, $\mathcal A$ is the forward operator that models the image acquisition as in \eqref{measurement_model-compact}, while $\mathcal T$ is a lifting operator that constructs a multi-fold Toeplitz structured matrix $\mathcal T(\hat {\boldsymbol\rho})$ from the entries of $\hat {\boldsymbol\rho}$.

\begin{figure*}[ht!]
\begin{minipage}{\textwidth}
\centering
\begin{tabular}{ccc}
\subfloat[][Construction of the Toeplitz matrix $\mathcal{T}(\boldsymbol{\widehat{\rho}})$ of dimension $|\Delta| \times |\Lambda|$.]{\includegraphics[width=0.5\textwidth]{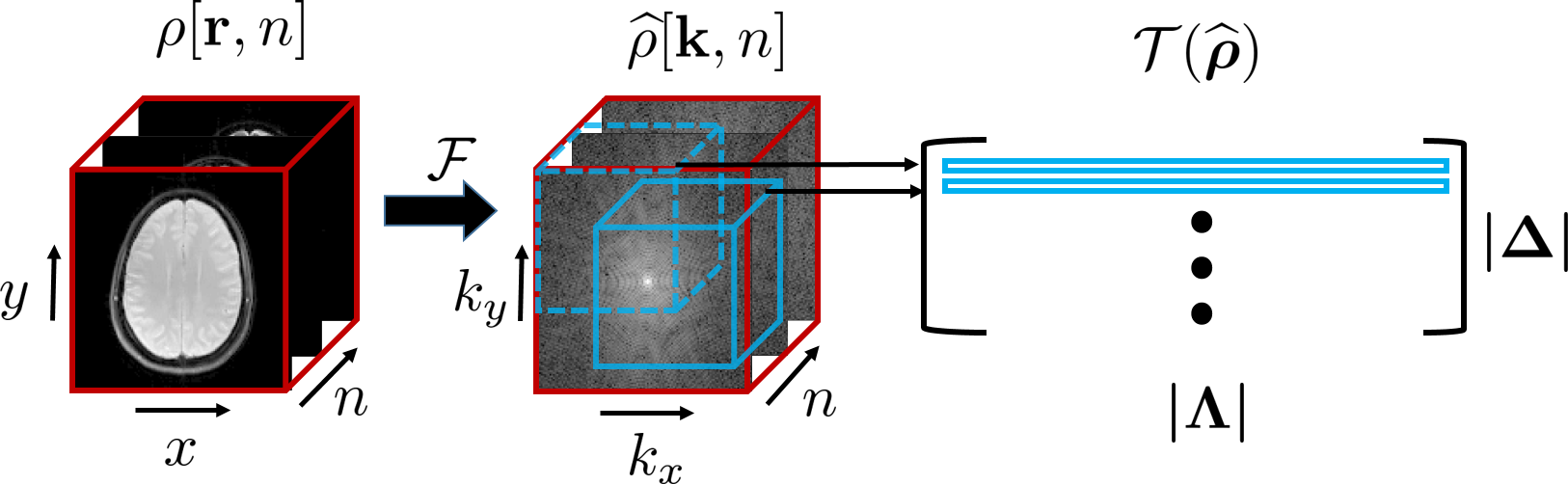}}&
\subfloat[][Illustration of the index sets for the construction of Toeplitz matrix.]{\includegraphics[ width=0.22\textwidth]{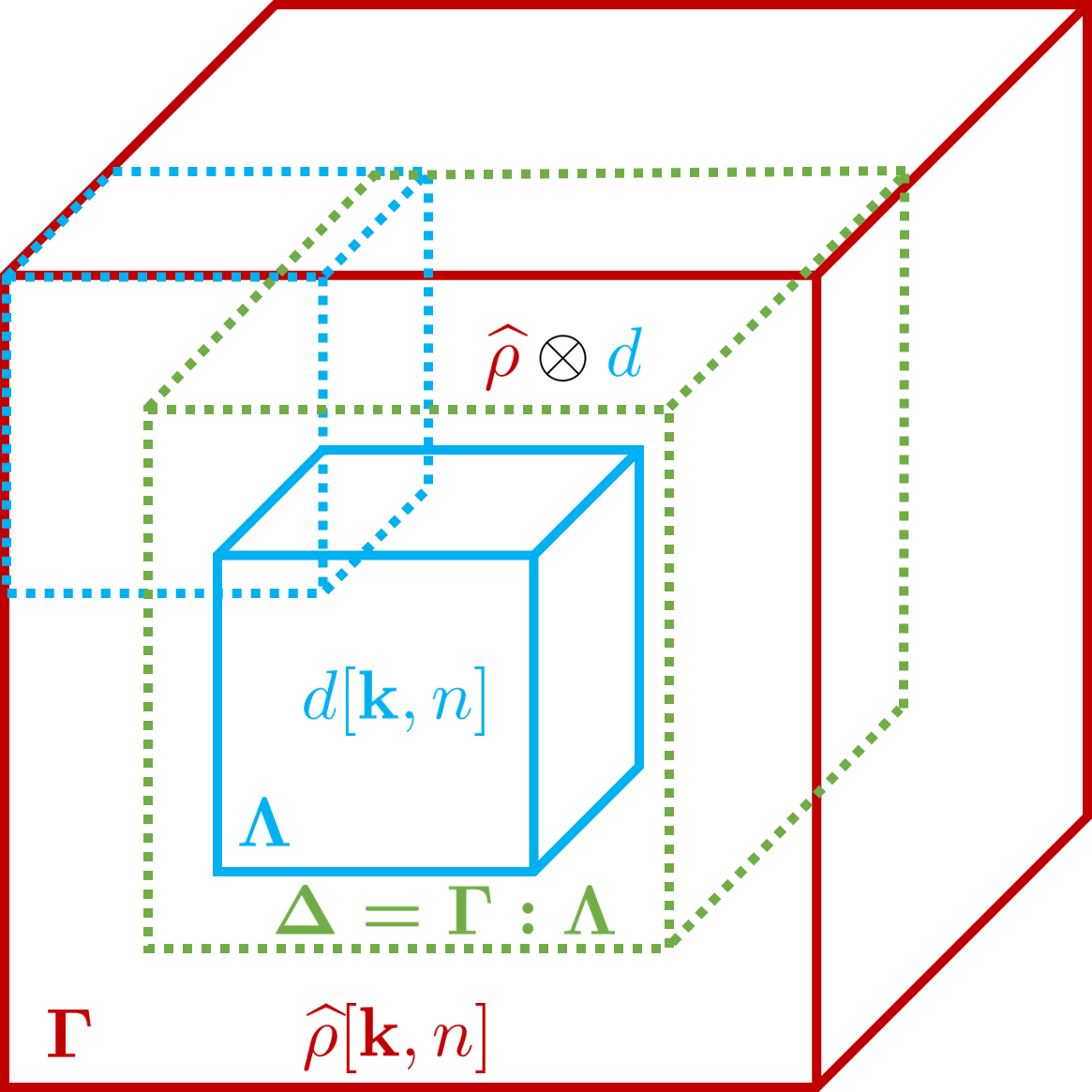}}&
\subfloat[][Illustration of the index sets for the minimal and assumed filters.]{\includegraphics[width=0.16\textwidth]{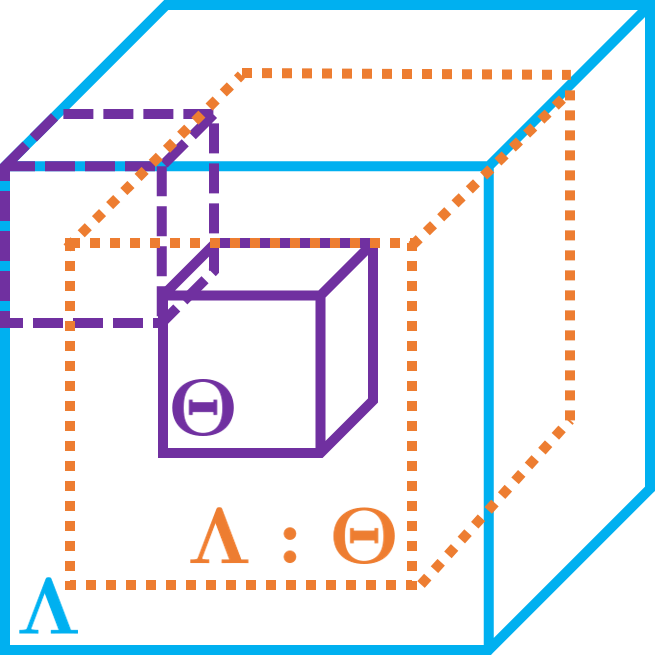}}
\end{tabular}
\end{minipage}
\caption{(a) Illustration of the construction of the matrix $\mathcal{T}(\boldsymbol{\widehat{\rho}})$ from the dataset $\boldsymbol{\widehat\rho}$: The rows of the Toeplitz matrix correspond to the cuboid shaped neighborhoods of the Fourier samples. The number of columns  is equal to the size of the filter support $(|\Lambda|)$. Similarly, the number of rows is equal to the number of valid linear convolutions between $\widehat{\rho}$ and the filter, denoted by $\Delta$. b) Illustration of the relation between the filter and signal supports and the matrix dimensions: The Fourier samples $\widehat{\rho}[\mathbf{k},n]$ and the filter coefficients $d[\mathbf{k},n]$ are assumed to be supported on the rectangular sets $\Gamma$ and $\Lambda$, respectively. The 3-D convolution between them is valid in the dotted rectangular region (in green) and the set of valid indices is represented by $\Delta = \Gamma:\Lambda$. c) The rectangular sets containing the coefficients of the minimal and assumed filters are represented by $\Theta$ and $\Lambda$ respectively. The number of linearly independent null space vectors of $\mathcal{T}(\boldsymbol{\widehat{\rho}})$ is given by all possible valid shifts of $\Theta$ in $\Lambda$, denoted by $\Lambda:\Theta$; this implies that $\mathcal{T}(\boldsymbol{\widehat{\rho}})$ is low rank and we enforce this property to estimate the missing entries of the matrix.
}\vspace{-1em}
\label{Fig1}
\end{figure*}

\section{Recovery using Annihilation Relations}

\subsection{Measurement model}
We consider the recovery of a series of images $\boldsymbol{\rho}$ from its multichannel Fourier measurements $\mathbf{b}$. The multichannel Fourier data $\boldsymbol{\widehat{\rho}}_{i}$ corresponding to the $i^{th}$ image frame can be modeled as
\begin{equation}
\label{measurement_ij}
\mathbf{b}_{ij} = \underbrace{\mathbf{S}_i \mathbf{F} \mathbf{C}_j \mathbf{F}^{*}}_{\mathbf{A}_{ij}} \boldsymbol{\widehat{\rho}}_{i}+ \boldsymbol{\eta}_{ij}, ~j = 1 \ldots \mbox{$N_{coils}$} 
\end{equation}
where $\mbf b_{ij}$ and $\boldsymbol{\eta}_{ij}$ are the under-sampled Fourier measurements and zero mean white gaussian noise corresponding to the $i^{th}$ frame and $j^{th}$ coil respectively, $\mbf C_j$ is the receiver coil sensitivity of the $j^{th}$ coil, $\mbf S_i$ is the sampling mask for the $i^{th}$ frame and $\mbf F$ is the 2D discrete Fourier transform (DFT) matrix. \eqref{measurement_ij} can be compactly written as
\begin{equation}
\label{measurement_model-compact}
\mathbf{b}= \mathcal{A}(\mathbf{\boldsymbol{\widehat{\rho}}}) + \boldsymbol{\eta}
\end{equation}
where $\boldsymbol{\widehat{\rho}} = \left[\boldsymbol{\widehat{\rho}}_1, \boldsymbol{\widehat{\rho}}_2, \ldots \boldsymbol{\widehat{\rho}}_T \right] \in \mathbb{C}^{B \times T}$ is the Fourier data in Casorati matrix form  \cite{liang2007spatiotemporal} with the $i^{th}$ column representing the vector of Fourier data at time instant $t_i$. $\mathcal{A}$ is a linear operator representing Fourier under-sampling and multiplication of coil sensitivities with $\boldsymbol{\widehat{\rho}}$.

\subsection{Annihilation property of smooth exponentials}
We model the signal at the spatial location $\mathbf{r} = (x,y)$ as a linear combination of $L$ exponentials:
\begin{equation}
\label{eq:signal model}
\rho[\mathbf{r},n]= \sum_{i=1}^{L}\alpha_{i}(\mathbf{r})~\beta_{i}(\mathbf{r})^{n},
\end{equation}
where  $\alpha_{i}(\mathbf{r}) \in \mathbb{C}$ are the amplitudes, $\beta_{i}(\mathbf{r}) \in \mathbb{C}$ is the exponential parameter that is dependent on the underlying physiology, $n$ refers to the signal index along the parameter dimension and $L$ is the number of exponentials at the voxel. For instance, in $T_2$ mapping applications the exponential parameters at the pixel location indexed by $\mathbf{r}$ are given by $\beta_i(\mathbf{r}) = \exp\left(\frac{-\Delta T}{T_{2,i}(\mathbf r)}\right)$. Here $\Delta T$ is the difference in echo times between two frames and $T_{2,i}$ is the relaxation parameter of the $i^{\rm th}$ tissue component (e.g. gray matter, CSF or white matter). 

The exponential signal, described in \eqref{eq:signal model}, at each pixel location can be annihilated by a 1-D FIR filter $g[\mathbf{r},n]$ \cite{stoica_spec}: 
\begin{equation}
\label{eq:annihilation 1D}
\sum_{m=0}^{L}\rho[\mathbf r,m]~g[\mathbf r,n-m] = 0, ~~ \forall \mbf r.
\end{equation}
where \eqref{eq:annihilation 1D} represents a 1-D convolution between the signal $\rho[\mathbf{r},n]$ and the $L+1$ tap filter $g[\mathbf{r},n]$. Since the exponential parameters vary from pixel to pixel, the filter $g[\mathbf r,n]$ also varies with the spatial location $\mbf r$. 

In practice, the exponential parameters vary smoothly as a function of space. This implies that the coefficients of the filter $g[\mathbf r,n]$  can be assumed to be smooth functions of the spatial variable $\mbf r$. Taking the 2-D Fourier transform of \eqref{eq:annihilation 1D} along the spatial dimensions, we obtain the following annihilation relation in the Fourier domain:
\begin{equation}
\label{eq:annihilation 3D}
\widehat\rho[\mathbf{k},n] \otimes d[\mathbf{k},n] = 0.
\end{equation} 
where $\widehat\rho[\mathbf{k},n] \stackrel{\mathcal F_{\rm 2D}}{\leftrightarrow} \rho[\mathbf r,n]$ and $d[\mathbf{k},n] \stackrel{\mathcal F_{\rm 2D}}{\leftrightarrow} g[\mathbf r,n]$ are the spatial Fourier coefficients of $\rho[\mathbf{r},n]$ and $g[\mathbf{r},n]$, respectively. Here, $\otimes$ denotes 3-D convolution. 

Since the filter coefficients of $g[\mathbf r,n]$ vary smoothly as a function of space, we assume $d[\mathbf{k},n]$ to be a 3-D FIR filter, whose coefficients are support limited in the rectangular set $\Lambda \subset \mathbb Z^3$; the size of $\Lambda$ (spatial bandwidth of $d[\mathbf{k},n]$) controls the spatial smoothness of the parameters, while the bandwidth along the parameter dimension is dependent on the number of exponentials in the signal model. 

We focus on the recovery of the Fourier coefficients of the signal specified by $\boldsymbol{\widehat\rho}$ within the rectangular set $\Gamma \subset \mathbb Z^3$. For simplicity, we assume $\boldsymbol{\widehat{\rho}}$ has $T$ frames, each of dimension $P \times Q$. The set $\Gamma$ is illustrated by the red cuboid in Fig. \ref{Fig1}.(b). The 3-D convolution \eqref{eq:annihilation 3D} can be compactly written as
\begin{equation}
\label{annihil-compact}
\mathcal{T}(\boldsymbol{\widehat{\rho}})\;\mathbf{d} =0
\end{equation}
where $\mathcal{T}$ is a linear operator that maps a 3-D dataset $\boldsymbol{\widehat{\rho}}$ into a lifted matrix $\mathcal{T}(\boldsymbol{\widehat{\rho}}) \in \mathbb{C}^{m \times s}$. The construction of the matrix is illustrated in Fig. \ref{Fig1}.(a). Similarly $\mathbf{d}$ represents the vectorized 3-D filter $d[\mathbf{k},n]$. Here, $s= |\Lambda|$ is the number of the columns of the matrix, where $\Lambda$ denotes the support of the filter $d$ indicated by the blue cuboid in Fig. \ref{Fig1}.(b). The number of rows in the matrix is denoted by $m = |\Gamma:\Lambda|$, which corresponds to the number of valid linear convolutions between $\boldsymbol{\widehat \rho}$ and the filter; the convolutions are valid in the green dotted cuboid in Fig. \ref{Fig1}.(b). The set $\Gamma:\Lambda$ is specified by
 \begin{equation}
 \Gamma:\Lambda = \{\mbf v \mid \Lambda + \mbf v \subseteq \Gamma; \mbf v \in \mbb Z^3\}.
\end{equation}
The blue dotted cuboid in Fig. \ref{Fig1}.(b) indicates a valid shift of the blue solid cuboid, whose support is given by $\Lambda$. We observe that $\mathcal{T}(\boldsymbol{\widehat{\rho}})$ has a multi-fold Toeplitz structure since the matrix-vector product in \eqref{annihil-compact} corresponds to a 3-D convolution.
%
\subsection{Dimensions of the fundamental subspaces of $\mathcal{T}(\widehat{\rho})$}
We denote the index set of the filter with the smallest support (termed as the minimal filter), which satisfies the annihilation relation, to be $
\Theta$. In practice, the support $\Theta$ (denoted by the purple cuboid in Fig. \ref{Fig1}.(c) is unknown. In such cases, the support set of the filter is overestimated to $\Lambda $, such that $\Theta \subset \Lambda$.  Let the dimensions of $\Lambda$ be $N_1\times N_2\times M$.
 
When the size of the filter is overestimated to $\Lambda$, it will result in multiple linearly independent vectors $\mbf d$ in the null space of $\mathcal{T}(\boldsymbol{\widehat{\rho}})$. Specifically, if $c[\mathbf{k},n]$ is the minimal filter, then any FIR filter of the form 
\begin{equation}
\label{eq:bigger_filter}
d[\mathbf{k},n] = c[\mathbf{k},n] \otimes e[\mathbf{k},n]
\end{equation}
 will also satisfy the following annihilation relation $\hat\rho[\mathbf{k},n] \otimes d[\mathbf{k},n] = 0$, or equivalently $\mathcal{T}(\boldsymbol{\widehat{\rho}})\mathbf{d} = 0$. Here, $e[\mathbf{k},n]$ is any FIR filter such that $d[\mathbf{k},n]$ is support limited to $\Lambda$. The number of such filters $(d[\mathbf{k},n])$ is specified by the set of all valid shifts of $\Theta$ in $\Lambda$, denoted by $\Lambda:\Theta$ \cite{gregsiam}; this set is indicated by the orange cuboid in Fig. \ref{Fig1}.(c). The corresponding shifted filters are linearly independent \cite{gregsiam}.

The above discussion shows that the dimension of the kernel of $\mathcal{T}(\boldsymbol{\widehat{\rho}})$ is at least $| \Lambda:\Theta|$. i.e.,
\begin{equation}
\dim\Big({\rm ker}\left(\mathcal T\left(\boldsymbol{\widehat{\rho}}\right)\right)\Big) \geq |\Lambda:\Theta|
\end{equation}
where $\Lambda$ is the assumed filter size and $| \Lambda:\Theta|$ denotes the cardinality of the set $ \Lambda:\Theta$. By the rank-nullity theorem, the rank of $\mathcal T\left(\boldsymbol{\widehat{\rho}}\right)$ or the dimension of the image space of $\mathcal T\left(\boldsymbol{\widehat{\rho}}\right)$ is specified by
\begin{equation}
{\rm rank}\Big(\mathcal T\left(\boldsymbol{\widehat{\rho}}\right)\Big)=\dim\Big({\rm im}\left(\mathcal T\left(\boldsymbol{\widehat{\rho}}\right)\right)\Big) \leq |\Lambda| - |\Lambda:\Theta|.
\end{equation}

Consider two datasets $\boldsymbol{\widehat{\rho}}_1$ and $\boldsymbol{\widehat{\rho}}_2$, where the size of $\Theta$ (purple cuboid in Fig. \ref{Fig1}.(c)) is smaller for $\boldsymbol{\widehat{\rho}}_2$. Note that a smaller minimal filter corresponds to a dataset with fewer exponentials and smoother parameters. Since the number of valid shifts indicated by the set $|\Lambda:\Theta|$ is higher for the dataset $\boldsymbol{\widehat{\rho}}_2$, we have ${\rm rank}(\mathcal{T}(\boldsymbol{\widehat{\rho}}_2)) < {\rm rank}(\mathcal{T}(\boldsymbol{\widehat{\rho}}_1))$. Hence, the rank of the Toeplitz matrix can be used as a measure of complexity of the dataset. 

Since we expect the matrix $\mathcal{T}(\boldsymbol{\widehat{\rho}})$ to be low rank, we use the rank prior as well as the Toeplitz structure of the  matrix to recover the exponential image series from under-sampled Fourier measurements.  The support of the different index sets along with the construction of the Toeplitz matrix is illustrated in Fig. \ref{Fig1}.

\subsection{Structured low-rank recovery from few measurements}

Since \eqref{measurement_model-compact} is an ill-posed problem, we employ the structured low rank matrix prior proposed in the previous sub-sections for the recovery of missing Fourier samples. We formulate the recovery of the Fourier data $\mathbf{\hat{\rho}}$ from the under-sampled measurements $\mathbf{b}$ as the following structured matrix completion problem:
\begin{equation}
\label{eq:problem formulation}
\min_{{\boldsymbol{\widehat{\rho}}}}~\text{rank}\left[\mathcal{T}(\boldsymbol{\widehat{\rho}})\right]~ \mbox{such that}~ \mathbf{b} = \mathcal{A}(\boldsymbol{\widehat{\rho}})+ \boldsymbol{\eta}
\end{equation}
Since the problem specified in \eqref{eq:problem formulation} is NP hard, we relax the rank function with a Schatten $p ~(0 \le p \le 1)$ norm. The relaxed objective function is then given by
\begin{equation}
\label{eq: problem formulation-relaxed}
{\boldsymbol{\widehat{\rho}}}^\star = \arg\min_{{\boldsymbol{\widehat{\rho}}}}  \|\mathcal{T}(\boldsymbol{\widehat{\rho}})\|_p + \frac{\mu}{2} \|\mathcal{A}(\boldsymbol{{\widehat{\rho}}}) - \mathbf{b}\|^2_2
\end{equation}
where $\mu$ is a regularization parameter that balances the weight given to the Schatten norm of the matrix and the data consistency term in \eqref{eq: problem formulation-relaxed}. $\mathcal{T}(\boldsymbol{\widehat{\rho}}) \in  \mathbb{C}^{m \times s}$ is a multifold Toeplitz matrix formed from the Fourier samples $\boldsymbol{\widehat{\rho}}$. $\|\mathbf{X}\|_p$ is the Schatten $p$ norm, defined as $\|\mathbf{X}\|_p : = \frac{1}{p}{\rm Tr}[(\mathbf{X}^H\mathbf{X})^{\frac{p}{2}}] = \frac{1}{p}{\rm Tr}[(\mathbf{X} \mathbf{X}^H)^{\frac{p}{2}}]=\frac{1}{p}\sum_i \sigma_i^p$; $\sigma_i$ are the singular values of $\mathbf{X}$. When $p=1$, the Schatten norm reduces to the convex nuclear norm and for $(0 \le p<1)$, the Schatten norm is a non-convex penalty; When $p\rightarrow0$, $\|\mathbf{X}\|_p : = \sum_{i} \log \sigma_i$. In section \ref{opt_alg}, we will focus on the algorithm to solve the optimization problem \eqref{eq: problem formulation-relaxed}.

\subsection{Relation to pixel-wise structured low-rank priors in \cite{MORASA}:}
The recovery of MR parameter weighted images considered in \cite{MORASA} is closely related to the proposed work. In \cite{MORASA}, the sum of structured low rank priors, formed at every pixel, is considered. Specifically, for every pixel, a Hankel matrix is constructed using the temporal signal at that pixel. The images are reconstructed by exploiting the low rank structure of all these matrices. The objective function is specified by
\begin{equation}
\label{eq: MORASA-objective}
\{\boldsymbol{\bar{\rho}}_{m}\} = \arg \min_{\boldsymbol{\bar{\rho}}} \sum_{i=1}^{l}\text{rank}[\mathcal{T}(\boldsymbol{\bar{\rho}}(\mathbf{r}_{i}))], ~\mbox{such that}~ \mathcal{A}(\boldsymbol{\bar{\rho}}) = \mathbf{b}
\end{equation}
 where $\boldsymbol{\bar{\rho}}_{m}$ is the set of images to be recovered, $l$ is the total number of pixels, $\mathcal{A}$ is a linear operator and $\mathbf{b}$ is a vector of measurements. 

We now consider a special case of our setting, where the spatial dimensions of $\Lambda$ (assumed filter size) are the same as that of the dataset (i.e. $P=N_1$ and $Q=N_2$), which is related to the above model. Since the spatial dimensions of the filter are the same as the dataset, no spatial smoothness is assumed on the annihilation filter coefficients. In this case, the dimension of the Toeplitz matrix $\mathcal{T}(\boldsymbol{\rho}) $, where $\boldsymbol{\rho}$ is a signal of interest, is specified by $(T-M+1) \times (N_{1}\cdot N_{2}\cdot M)$. The Toeplitz matrix after a re-arrangement of the columns has the following structure:
\begin{equation}
\label{eq:Toeplitz_g}
\mathcal{T}(\boldsymbol{\rho}) = \Big(\begin{array}{c|c|c|c}
\mathcal{T}(\boldsymbol{\rho\big(\mathbf{r}}_{1})\big) & \mathcal{T}\big(\boldsymbol{\rho(\mathbf{r}}_{2})\big)& \ldots & \mathcal{T}\big(\boldsymbol{\rho(\mathbf{r}}_{l})\big)   
\end{array}\Big)
\end{equation}
In \eqref{eq:Toeplitz_g}, each of the Toeplitz matrices $\mathcal{T}(\boldsymbol{\rho({\mathbf{r}}}_{i})) \in \mathbb{C}^{(T-M+1)\times M}$, whose entries correspond to the temporal signal at location $\mathbf{r}_{i}$. We observe that enforcing a low rank prior on the special case considered in \eqref{eq:Toeplitz_g} results in a more constrained approach than the pixel-wise low rank penalty in \eqref{eq: MORASA-objective}. In particular, the global low-rank prior considered in \eqref{eq:Toeplitz_g} enables the exploitation of correlations between the columns of $\mathcal{T}(\boldsymbol{\rho})$, in addition to the annihilation relations. In contrast, the pixel-wise approach in \eqref{eq: MORASA-objective} is not capable of exploiting these correlations. An additional low-rank prior on the Casorati matrix of the images or a wavelet prior has to be used as in \cite{MORASA} to exploit these correlations. Since the special case of our formulation is already capable of exploiting these correlations, we do not require additional priors. More importantly, we consider spatially bandlimited filters, which account for the smoothness of the exponential parameters. We observe that this property offers a 3 dB improvement (see Table \ref{tab:table1}) over the special case considered in \eqref{eq:Toeplitz_g}. In summary, the structured low-rank prior considered in this work qualitatively consolidates and unifies the multiple diverse priors used in \cite{MORASA}. 

Another key benefit of the above special case is the reduced computational complexity. When an iterative re-weighted least squares (IRLS) based approach \cite{irls} is employed, each step of the algorithm \eqref{eq: MORASA-objective} requires the eigen decomposition of as many Gram matrices as pixels in the dataset. In contrast, the use of the prior \eqref{eq:Toeplitz_g} requires the eigen decomposition of only one Gram matrix at each iteration. As the dimension of the Gram matrices in both the cases is the same and equal to $(T-M+1) \times (T-M+1)$, the computational complexity of the special case is orders of magnitude lower than that of the pixel-wise structured low rank strategy. 

\section{Optimization Algorithm}
\label{opt_alg}
The minimization of \eqref{eq: problem formulation-relaxed} using classical low-rank matrix recovery schemes is challenging due to the large size of the structured matrix, which often requires several orders of magnitude more memory, when compared to the original 3-D dataset. Current structured low-rank methods \cite{loraks,ALOHA} employ approaches originally designed for low-rank matrix recovery and do not exploit the structure of the matrix; the direct use of these 2-D algorithms to our 3-D setting is difficult due to the large memory demand and computational complexity. 

We modify the GIRAF algorithm \cite{GIRAF} to the 3-D setting to minimize the computational complexity. Specifically, we employ an IRLS based algorithm \cite{irls} to solve the optimization problem in \eqref{eq: problem formulation-relaxed}. This approach allows us to use efficient approximations for operations involving the Toeplitz matrix using fast Fourier transforms; these modifications quite significantly reduce the computational complexity and memory demand. To derive the basic idea of the algorithm, we use the following identity to express the Schatten $p$ norm of a matrix as a weighted Frobenius norm:
\begin{align}
\label{eq: irls_identity}
\|\mathbf{Y}\|_p & =\frac{1}{p}{\rm Tr}[\underbrace{(\mathbf{Y} \mathbf{Y}^*)^{\frac{p}{2}-1}}_{\mathbf{H}}\mathbf{Y}\mathbf{Y}^*] \\
\label{eq: irls_identity_simplified}
& =  \frac{1}{p}\|\mathbf{H}^{\frac{1}{2}}\mathbf{Y}\|^{2}_F
\end{align} 
Hence the solution to the minimum Schatten norm can be obtained by alternating between the update of a weight matrix $\mathbf{H}$ and the solution to a weighted least squares problem. In our case, we set $\mathbf{Y} = \mathcal{T}(\boldsymbol{\widehat{\rho}})$ in \eqref{eq: irls_identity}, which decouples \eqref{eq: problem formulation-relaxed} into two sub-problems. At the $n^{th}$ iteration, the sub-problems are given by
\begin{eqnarray}\nonumber
\label{eq:x-update}
{\boldsymbol{\widehat{\rho}}}^{(n)} & =& \arg \min_{{\boldsymbol{\widehat{\rho}}}}\| (\mathbf{\sqrt{H}})^{(n-1)}\,\mathcal{T}(\boldsymbol{\widehat{\rho}})\|_F^2 + \frac{\mu p}{2}\|\mathcal A(\boldsymbol{\widehat{\rho}}) -\mathbf{b}\|_2^2\\\\
\label{eq:weight-update}
\mathbf{H}^{(n)} & =& [\underbrace{\mathcal{T}(\boldsymbol{\widehat{\rho}}^{(n)})\,\mathcal{T}(\boldsymbol{\widehat{\rho}}^{(n)})^*}_{\mathbf{R}} + \epsilon^{(n)} \,\mathbf{I}]^{\frac{p}{2}-1}
\end{eqnarray}
where $\epsilon^{(n)} \rightarrow 0$ is added to stabilize the inverse.  Hence to solve \eqref{eq: problem formulation-relaxed}, we employ an alternating minimization scheme that cycles between the sub-problems  \eqref{eq:x-update} and \eqref{eq:weight-update} till the cost of \eqref{eq: problem formulation-relaxed} between successive iterates is below a tolerance threshold. In the next two sub-sections, we describe an efficient implementation of the two sub-problems.

\subsection{Least squares solution of \eqref{eq:x-update}}
Let the rows of $\mathbf{\sqrt{H}}$ be denoted by $\left[(\mathbf{h}^{(1)})^{T},\ldots,(\mathbf{h}^{(M)})^{T}\right]^{T}$. Substituting for $\mathbf{\sqrt{H}}$ in \eqref{eq:x-update}, we obtain
\begin{equation}
\label{eq:ls_update}
{\boldsymbol{\widehat{\rho}}}^* = \arg \min_{{\boldsymbol{\widehat{\rho}}}}\sum_{i=1}^{M}\|\mathbf{h}^{(i)}\,\mathcal{T}(\boldsymbol{\widehat{\rho}})\|^{2}_{2}~+~\frac{\mu p}{2}\,\|\mathcal A(\boldsymbol{\widehat{\rho}}) -\mathbf{b}\|_2^2
\end{equation}
The term $\mathbf{h}^{(i)}\,\mathcal{T}(\boldsymbol{\widehat{\rho}})$ in \eqref{eq:ls_update} represents a 3-D linear convolution between the 3-D sequences $\mathbf h^{(i)}$ and $\boldsymbol{\widehat\rho}$. 

In the GIRAF algorithm \cite{GIRAF}, the linear convolutions were approximated by circular convolutions, so that they could be efficiently implemented using Fast Fourier transforms (FFTs). The approximations were valid due to the rapid decay of the Fourier coefficients towards the boundaries. However, in our case as the  signal at every voxel follows an exponential curve, the magnitude of the Fourier coefficients are high at the first few points along the parameter dimension. Hence, the direct application of the GIRAF scheme to our setting gives poor results. 

We introduce a hybrid strategy to improve the approximations, while keeping the memory demand and computational complexity low. In particular, we approximate the 3-D linear convolution as a series of 2-D circular convolutions along the spatial dimensions and a linear convolution along the parameter dimension. Denoting the $l^{\rm th}$ frame of $\boldsymbol{\widehat \rho}$ and $h$ as $\widehat \rho_{l}\left[\mbf k\right]$ and $h_l\left[\mbf k\right]$ respectively, we rewrite the convolution relation as 
\begin{eqnarray}\nonumber
h[\mathbf{k},n] \otimes \mathbf{\widehat{\rho}}[\mathbf{k},n]&=&\sum_{m}  \sum_{\mbf p}\widehat \rho\left[\mbf k-\mbf p, n-m\right] h[\mbf p,m]\\
&=& \sum_{m} \underbrace{ \sum_{\mbf p}\widehat \rho_{n-m}\left[\mbf k-\mbf p \right] h_m[\mbf p]}_{g_{n-m,m}=\widehat\rho_{n-m} ~* ~h_m},
\end{eqnarray}
where $\otimes$ denotes 3-D convolution and $*$ denotes 2-D convolution. Since the spatial Fourier coefficients of $\widehat \rho\left[\mbf k\right]$ decay rapidly towards the boundaries, the 2-D linear convolutions $g_{j,l}=\mathbf{\hat\rho}_{j} ~* ~h_l$ can be approximated as 2-D circular convolutions and efficiently computed using fast Fourier transforms as shown in \cite{GIRAF}. After the 2-D convolutions are evaluated for all feasible combinations $g_{j,l}$, we compute the outer sum. 

Now, we express the aforementioned idea in compact matrix notations.  Let $\mathbf{h}$ and $\mathbf{\hat{\rho}}$ consist of $M$ and $T$ frames respectively. We consider an arbitrary filter $h$ of spatial dimensions $N_1 \times N_2$ and denote its $i^{\rm th}$ frame by $\mathbf{h}_i$. Now $(\mathbf{h}\; \mathcal{T}(\boldsymbol{\widehat{\rho}}))$ can be expanded as,
\begin{equation}
\label{eq:expand first term}
\mathbf{h} \;\mathcal{T}(\boldsymbol{\widehat{\rho}}) = \begin{pmatrix}
\mathbf h_{M}& \ldots & \mathbf h_{1}
\end{pmatrix}\begin{pmatrix}
 \mathbf{T}(\boldsymbol{\widehat{\rho}}_{1})& .. &\mathbf{T}(\boldsymbol{\widehat{\rho}}_{T-M+1})\\
\vdots & \vdots & \vdots \\
\mathbf{T}(\boldsymbol{\widehat{\rho}}_{M}) & .. &\mathbf{T}(\boldsymbol{\widehat{\rho}}_{T}) \\
\end{pmatrix}
\end{equation}
In the above equation, $\mathbf{T}(\boldsymbol{\widehat{\rho}}_{j})$ represents a Toeplitz matrix formed from the samples of $\boldsymbol{\widehat{\rho}}_j$. This matrix can be expressed in terms of a larger circulant matrix \cite{GIRAF} in the following way:
\begin{equation}
\label{Toeplitz - circ_rep}
\mathbf{T}(\boldsymbol{\widehat{\rho}}_j) \approx \mathbf{P}_{\Lambda_{s}}^*~ \mathbf{C}(\boldsymbol{\widehat{\rho}}_{j})
\end{equation}
Here $\mathbf{C}(\boldsymbol{\widehat{\rho}}_{j}) \in \mathbb{C}^{L \times L}$ is a circulant matrix formed from the Fourier samples $\boldsymbol{\widehat{\rho}}_j$, $\Lambda_{s}$ is the support of a frame of the filter and $\mathbf{P}_{\Lambda_{s}}^* \in \mathbb{C}^{N_{1}N_{2} \times L}$ corresponds to zero padding operation outside the filter support $\Lambda_{s}$. Note that the support of $\mathbf h_l$ is often much smaller than that of $\boldsymbol{\widehat \rho}_j$.

Using the approximation in \eqref{Toeplitz - circ_rep}, we can efficiently evaluate $\mathbf{h}_{l}\;\mathbf{T}(\boldsymbol{\widehat{\rho}}_j)$, which is the 2-D linear convolution between the $l^{th}$ frame of $\mathbf{h}$ and $j^{th}$ frame of $\boldsymbol{\widehat{\rho}}$, as
\begin{eqnarray}\nonumber
\mathbf{h}_{l}\;\mathbf{T}(\boldsymbol{\widehat{\rho}}_j) &\approx& \mathbf{h}_{l}\;\mathbf{P}_{\Lambda_s}^{*} \;\mathbf{C}(\boldsymbol{\widehat{\rho}}_{j})\\
\label{eq:circ_rep}
&=&\left[\boldsymbol{\widehat{\rho}}_{j}\right]^{T}\underbrace{\mathbf{C}(\mathbf{h}_{l}\;\mathbf{P}_{\Lambda_{s}}^{*})}_{\mathbf{C}_{l}} .
\end{eqnarray}
where we have used the commutative property of convolution to arrive at the expression in \eqref{eq:circ_rep}. Here, $\mathbf{C}_{l} =\mathbf{C}(\mathbf{h}_{l}\mathbf{P}_{\Lambda_s}^{*} )$ is a circulant matrix formed from the zero padded filter coefficients $\mathbf{h}_{l}$. Hence, the product $\left[\boldsymbol{\widehat{\rho}}_{j}\right]^{T}\mathbf{C}_{l}$ denotes the 2-D circular convolution between $\left[\boldsymbol{\widehat{\rho}}_{j}\right]^{T}$ and the zero-padded filter coefficients. 

We propose to implement the circular convolutions using fast Fourier transforms to minimize the computational complexity. Specifically, we compute \eqref{eq:circ_rep} efficiently as 
\begin{eqnarray}
\label{eq:circ_fft}
\left[\boldsymbol{\widehat{\rho}}_{j}\right]^{T}\mathbf{C}_{l} &=& \left[\boldsymbol{\widehat{\rho}}_{j}\right]^{T}~\underbrace{\mathbf{F}^* \mathbf D_{l}\mathbf{F}}_{\mathbf{C}_{l}}
\end{eqnarray}
where $\mathbf D_l$ is a diagonal matrix with diagonal entries $\mu_l \stackrel{\mathcal F_{2D}}{\leftrightarrow} \mathbf{h}_{l}\;\mathbf{P}_{\Lambda_s}^{*}$ and $\mathbf F$ denotes the 2-D discrete Fourier transform matrix.

Using \eqref{eq:circ_fft}, we simplify \eqref{eq:expand first term} as,
\begin{equation}\nonumber\small
\label{eq:simplify first term}
\mathbf{h}\mathbf{T}(\boldsymbol{\widehat{\rho}})\approx 
\boldsymbol{\left[\widehat{\rho}\right]}^{T}\mathbf{Q}^* \underbrace{\begin{pmatrix}
\mathbf{D}_{M} &  \ldots && \mathbf{0}\\
\mathbf{D}_{M-1} & \mathbf{D}_{M} & &\vdots\\
\vdots& \vdots & \ddots\\
\mathbf{D}_1 & \cdots && \mathbf{0}\\
\mathbf{0} & \mathbf{D}_1 & &\mathbf D_{M}\\
\vdots&\vdots & \cdots&\vdots\\
\mathbf{0}&\mathbf{0}& \cdots & \mathbf{D}_{1}
\end{pmatrix}}_{\left[\mathbf{D}(\mathbf{h})\right]^{T}} \mathbf Q
\end{equation} 
\begin{figure}[t!]
\centering
\includegraphics[width=0.4\textwidth]{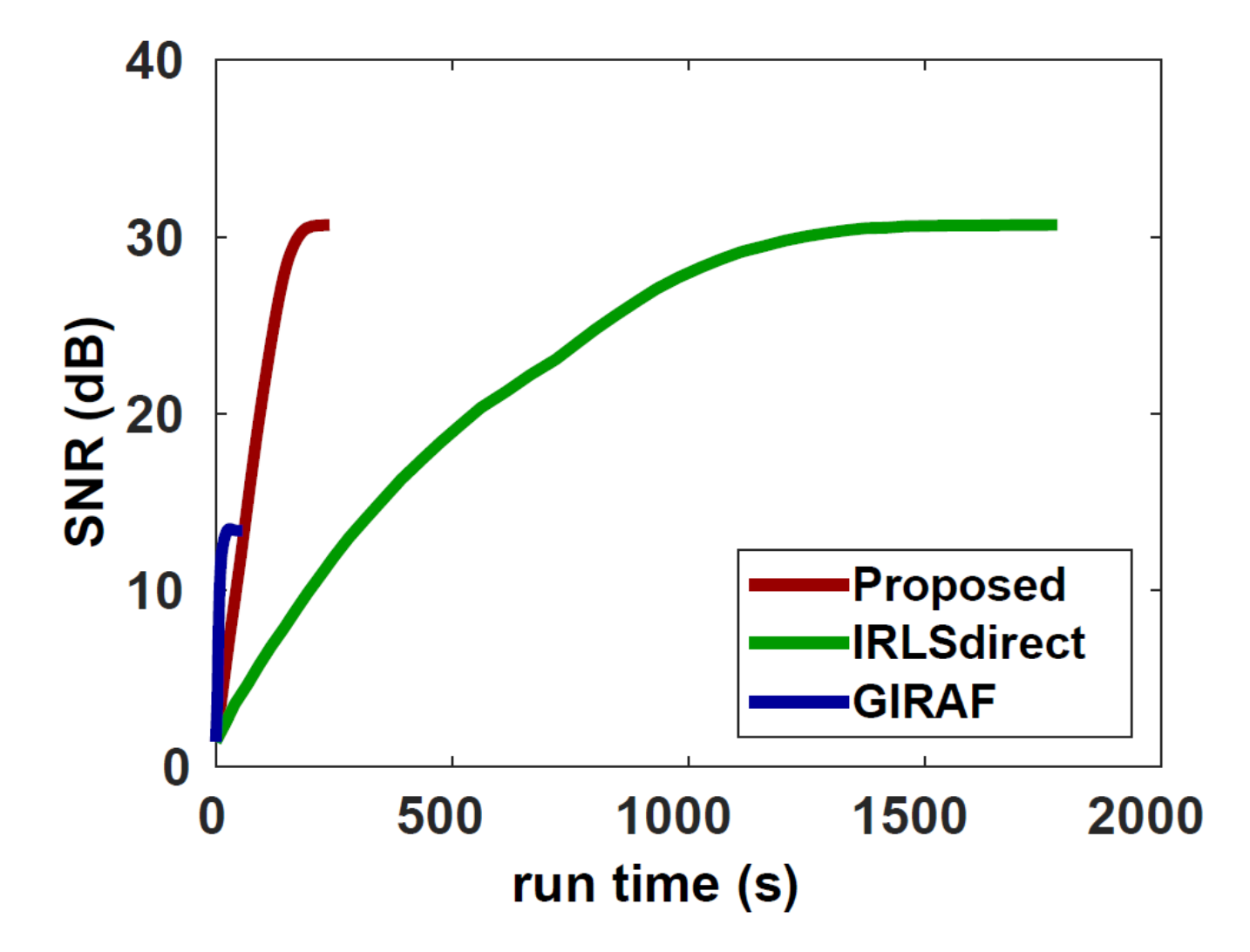}
\caption{Effect of approximations on the run time of the proposed algorithm and comparison of SNR: The approximations introduced in the proposed method enable efficient computation of the sub-problems \eqref{eq:x-update} and \eqref{eq:weight-update}  using fast Fourier transforms (FFTs). This results in a faster convergence (7.5 fold speed up) to the same solution as the one obtained using the IRLS (direct) method. }
\label{Fig2}
\end{figure}
where each of the matrices $\mathbf D_l$ are diagonal matrices and the dimension of $\mathbf{D}(\mathbf{h})^{T}$ is $PQT \times PQk$ with $k = T-M+1$. Here, ${\boldsymbol{[\widehat{\rho}}]}^T = \left[\boldsymbol{\widehat{\rho}}_{1}, \ldots, \boldsymbol{\widehat{\rho}}_T\right]^{T}$ and $\mathbf{Q}^* = \mathbf I\bigotimes \mathbf F^*$ is a block diagonal matrix with the diagonal blocks being the inverse 2-D Fourier transform matrix. Thus, we can express $\|\mathbf{h}\;\mathbf{T}(\boldsymbol{\widehat{\rho}})\|^2$ as $\|\mathbf{D}(\mathbf{h}) \mathbf{Q}^*\;\boldsymbol{\widehat{\rho}}\|^2$. 



Using the above relation and substituting for the first term in \eqref{eq:ls_update} we obtain,
 \begin{equation}
\label{ls_update_mod}
{\boldsymbol{\widehat{\rho}}}^* = \arg \min_{{\boldsymbol{\widehat{\rho}}}}\sum_{i=1}^{M}\|\mathbf{D}(\mathbf{h}^{(i)}) \mathbf{Q}^* \boldsymbol{\widehat{\rho}}\|^{2}_{2}+\frac{\mu p}{2}\|\mathcal A(\boldsymbol{\widehat{\rho}}) -\mathbf{b}\|_2^2
\end{equation}
The above equation can be solved by taking its gradient and setting it to zero. The gradient is given by
\begin{equation}
\label{eq:grad_ls}
2\mathbf{Q}\bigg(\underbrace{\sum_{i=1}^{M}(\mathbf{D}(\mathbf{h}^{(i)}))^{*}\mathbf{D}(\mathbf{h}^{(i)})}_{\mathbf{G} }\bigg)  \mathbf{Q}^{*}\boldsymbol{\widehat{\rho}}+  \mu p~ \mathcal A^* \mathcal A \boldsymbol{\widehat{\rho}}= \mu p ~\mathcal A^*\mathbf{b}
\end{equation}
Note that prior to solving \eqref{eq:grad_ls}, $\mathbf{G}$ can be precomputed efficiently. Denote $\mathbf{E}^{*} = \left[\mathbf{D}_M, \mathbf{D}_{M-1} \ldots, \mathbf{D}_1\right]$. In order to populate the entries of $(\mathbf{D}(\mathbf{h}^{(i)}))^{*}\mathbf{D}(\mathbf{h}^{(i)})$, we need to compute one product $(\mathbf{E}\mathbf{E}^*)^{(i)}$, and the sum of $k=(T-M+1)$ sparse matrices. Each sparse matrix contains a shifted version of $(\mathbf{E}\mathbf{E}^*)^{(i)}$ as the only non-zero block. After precomputing $\mathbf{G}$, we only need a few iterations of conjugate gradient (cg) algorithm to solve \eqref{eq:grad_ls}.

\begin{figure*}[t!]
\centering
\includegraphics[width=0.73\textwidth]{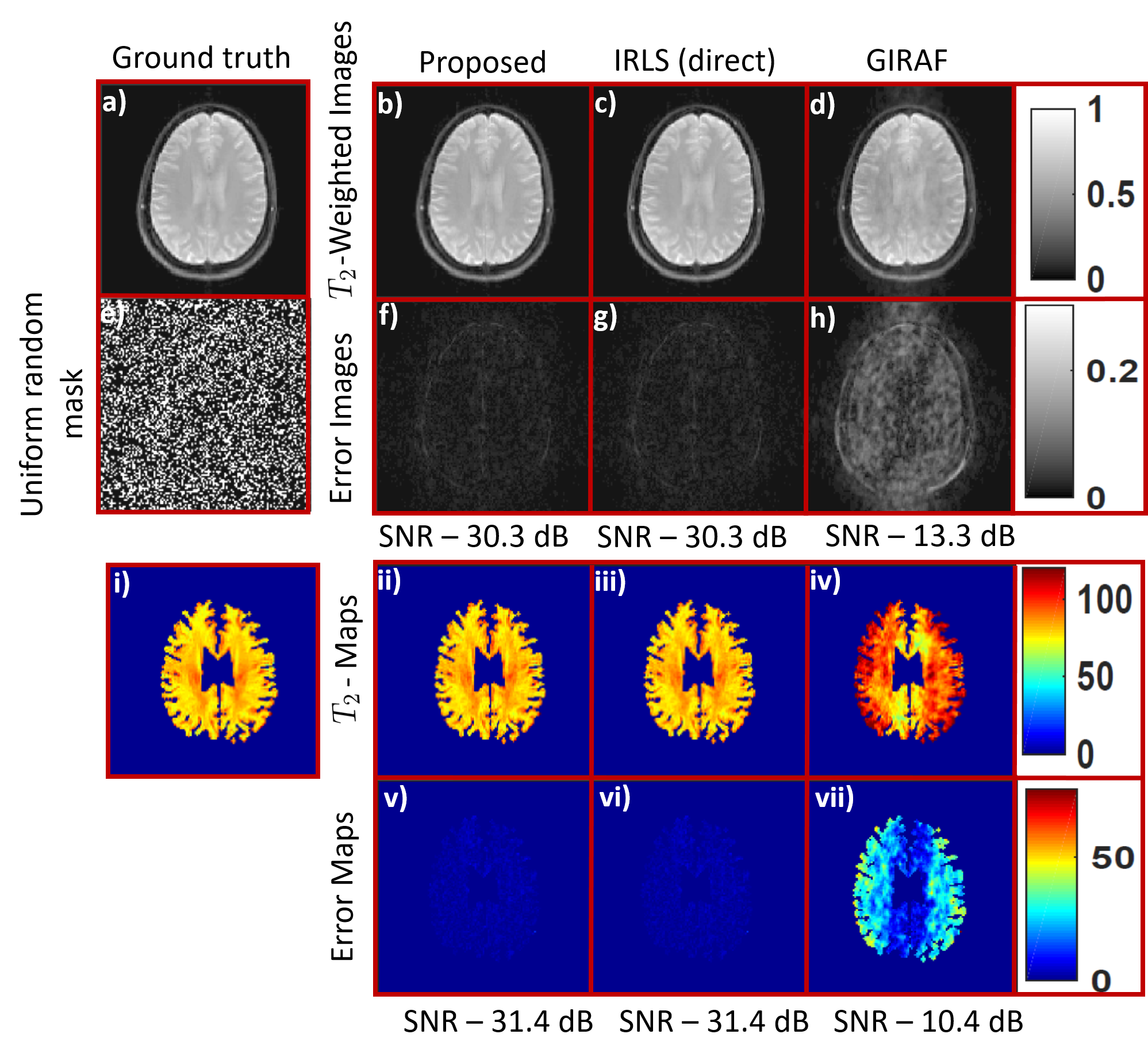}\vspace{-1em}
\caption{Effect of approximations introduced in the proposed method on the recovery of coil combined $T_2$ weighted images from 30 percent uniform random Fourier measurements: One frame (4\rm th Echo, TE = 40ms) of the image sequence corresponding to the ground truth is shown in (a) along with the frame of the sampling mask in (e). The 4\rm th echo of the reconstructed image sequence using proposed method (b) is compared with those obtained using IRLS (direct) in (c) and GIRAF in (d). The corresponding error images are shown in (f)-(h). The estimated $T_2$ maps, which were multiplied by a mask to remove the background and the CSF region, are shown in (i)-(iv) and the corresponding error maps are shown in (v)-(vii).}
\label{Fig3}
\end{figure*}

\begin{figure}[t!]
 \removelatexerror
  \begin{algorithm}[H]
   \caption{Proposed algorithm for the recovery of exponential image time series}
   Initialize $\boldsymbol{\widehat{\rho}}^{(0)}$ and choose $\epsilon^{(0)} > 0$\;
   \For{$n=1$ to $N_{\text{max}}$ or tolerance threshold reached}
   {
   		\textbf{Step1: Weight matrix update:}\\
   		\quad Compute each block $\mathbf{R}_{p,q}$ of the Gram matrix $\mathbf{R}$ using \eqref{eq:block-expression for R} and \eqref{eq:P_ij modified}\;
   		\quad Compute the eigen values and eigen vectors $\{\lambda^{(i)}, \mathbf{u}^{(i)}\}_{i=1}^{M}$ from the eigen decomposition of $\mathbf{R}$\;
   		\quad Evaluate the null space vectors:\\
   		\quad\quad$\mathbf{h}^{(i)} = \sqrt{\alpha_{i}}(\mathbf{u}^{(i)})^*$, where  $\alpha^{(i)} = (\mathbf{\lambda}^{(i)} + \epsilon^{(n-1)})^{\frac{p}{2}-1}$\;
   		\textbf{Step2: Least Squares update:}\\
   		\quad Solve the least squares problem:\\
   		\quad\quad ${\boldsymbol{\widehat{\rho}}}^{(n)} = \arg \min_{{\boldsymbol{\widehat{\rho}}}}\sum_{i=1}^{M}\|\mathbf{D}(\mathbf{h}^{(i)})\mathbf{Q}^* \boldsymbol{\widehat{\rho}}\|^{2}_{2}+\frac{\mu p}{2}\|\mathcal A(\boldsymbol{\widehat{\rho}}) -\mathbf{b}\|_2^2$\\
   		\quad using Conjugate gradient (CG) method\;
        Choose $\epsilon^{(n)}$ such that $0 < \epsilon^{(n)} \leq \epsilon^{(n-1)}$\;
   }		
  \label{alg:proposed_algo}
  \end{algorithm}
\end{figure}

\subsection{Weight-Update}
The first step in computing the weight matrix $\mathbf{H}^{\frac{1}{2}}$ involves forming the Gram matrix $\mathbf{R}=\mathcal{T}(\boldsymbol{\widehat{\rho}})\,\mathcal{T}(\boldsymbol{\widehat{\rho}})^*$. The direct computation of $\mathbf R$ requires the evaluation and storage of the lifted matrix $\mathcal{T}(\boldsymbol{\widehat{\rho}})$, which will be a computationally expensive and memory intensive operation. Instead we propose an efficient way to compute the Gram matrix. Specifically, we assume $\mathbf{R}$ to be partitioned in the following way:
\begin{equation}
\label{eq:gram matrix}
\begin{pmatrix}
\mathbf{R}_{1,1} & \mathbf{R}_{1,2} & \ldots & \mathbf{R}_{1,M}  \\
\mathbf{R}_{2,1} & \mathbf{R}_{2,2} & \ldots & \mathbf{R}_{2,M}  \\
\vdots & \vdots & \cdots & \vdots  \\
\vdots & \vdots & \cdots & \vdots  \\
\mathbf{R}_{M,1} & \mathbf{R}_{M,2} & \cdots & \mathbf{R}_{M,M}  \\
\end{pmatrix}
\end{equation}
where the above matrix has $M$ column and row partitions and $\mathbf{R}_{i,j}$ is a matrix block of dimension $N_{1} N_{2} \times N_{1} N_{2}$. We obtain a general expression for the matrix block corresponding to the $p^{th}$ row and $q^{th}$ column partition of $\mathbf{R}$ as
\begin{equation}
\label{eq:block-expression for R}
\mathbf{R}_{p,q} = \sum_{i=1}^{k}\mathbf{T}(\boldsymbol{\widehat{\rho}}_{p+i-1}) \mathbf{T}(\boldsymbol{\widehat{\rho}}_{q+i-1})^*
\end{equation}
where $k:=T- M+1$. To compute $\mathbf{R}_{p,q}$, we use the relation in \eqref{Toeplitz - circ_rep} and simplify $\mathbf{T}(\boldsymbol{\widehat{\rho}}_i)\mathbf{T}(\boldsymbol{\widehat{\rho}}_j)^*$ as
\begin{equation}
\label{eq:P_ij modified}
\mathbf{P}_{i,j} = \mathbf{T}(\boldsymbol{\widehat{\rho}}_i)\mathbf{T}(\boldsymbol{\widehat{\rho}}_j)^* =  \mathbf{P}_{\Lambda_{s}}^*\underbrace{\mathbf{C}(\boldsymbol{\widehat{\rho}}_{i})\mathbf{C}(\boldsymbol{\widehat{\rho}}_{j})^*}_{\mathbf{C}(\mathbf{g})}\mathbf{P}_{\Lambda_{s}}
\end{equation}
where the entries of $\mathbf{C}(\mathbf{g})$ are obtained from the array $\mathbf{g}$. The entries of $\mathbf{g}$ are given by $\mathbf{F}(\boldsymbol{\rho}_i \circ \mbox{conj}(\boldsymbol{\rho}_j))$, where $\boldsymbol{\rho}_{i}$ and $\boldsymbol{\rho}_{j}$ are the images corresponding to the Fourier samples $\boldsymbol{\widehat{\rho}}_i$ and $\boldsymbol{\widehat{\rho}}_j$ respectively, $\mathbf{F}$ denotes a 2-D DFT matrix, $\mbox{conj}$ denotes the conjugate operation and $\circ$ denotes point-wise multiplication. Hence the entries of every row of $\mathbf{P}_{i,j}$ can be populated by performing a sliding window operation that extracts and vectorizes a $N_{1} \times N_{2}$ patch from a $(2N_{1}-1)$ $\times$ $(2N_{2}-1)$ neighborhood. 

Next the weight matrix $\mathbf{H}^{\frac{1}{2}}$ is efficiently computed from the eigen decomposition of $\mathbf{R}$. Let $\mathbf{U}$ represent the orthogonal basis of eigen vectors $\mathbf{u}^{(i)}$ and $\Lambda$ be a diagonal matrix containing the eigen values $\lambda^{(i)}$. Then the eigen decomposition of $\mathbf{R}$ is given by $ \mathbf{U}\Lambda\mathbf{U}^*$. Substituting for $\mathbf{R}$ in (\ref{eq:weight-update}) and simplifying further we obtain,  
\[
\mathbf{H} = [\mathbf{U} (\mathbf{\Lambda}+\epsilon \mathbf{I})\mathbf{ U}^*]^{\frac{p}{2}-1} = \mathbf{U} (\mathbf{\Lambda}+\epsilon \mathbf{I})^{{\frac{p}{2}-1}} \mathbf{U}^*.
\] 
Hence, one choice of the matrix square root $\mathbf{H}^{\frac{1}{2}}$ is
\[
\mathbf{H}^{\frac{1}{2}} = (\mathbf{\Lambda}+\epsilon \mathbf{I})^{\frac{p}{4}-\frac{1}{2}}\mathbf{U}^* = \left[(\mathbf{h}^{(1)})^{T},\ldots,(\mathbf{h}^{(M)})^{T}\right]^{T}
\] 
where $\mathbf{h}^{(i)} = \sqrt{\alpha^{(i)}}(\mathbf{u}^{(i)})^*$ and $\alpha^{(i)} = (\mathbf{\lambda}^{(i)} + \epsilon)^{\frac{p}{2}-1}$.

\begin{figure*}[t!]
\centering
\includegraphics[width=0.75\textwidth]{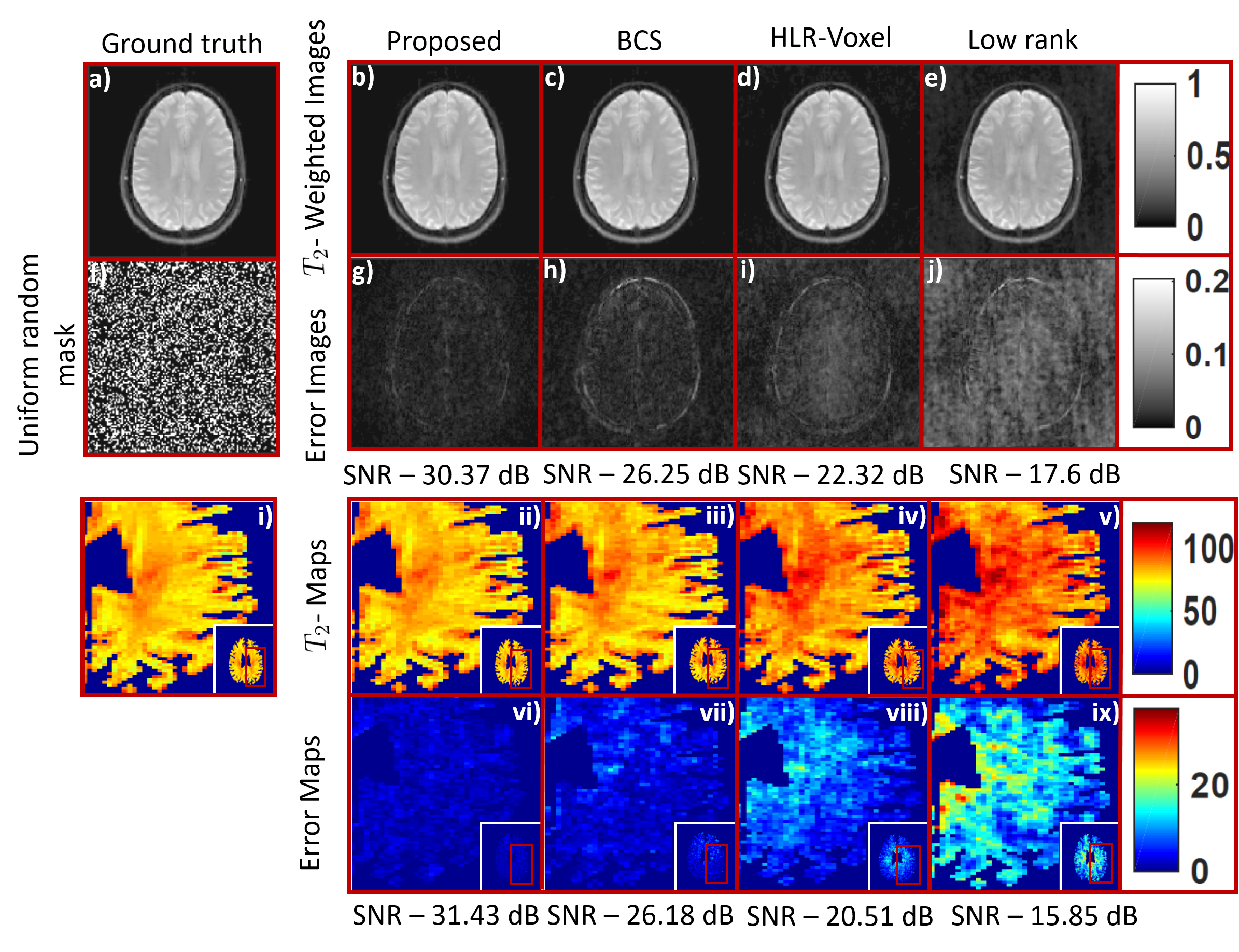}
\caption{Comparison of the proposed method with different reconstruction schemes on the recovery of coil combined data from 30 percent uniform random Fourier measurements: One frame (4\rm th Echo, TE = 40ms) of the image sequence is shown in (a)-(e) along with the frame of sampling mask in (f). The corresponding error images are shown in (g)-(j). The estimated $T_2$ maps, specifically the enclosed rectangular region is zoomed and is shown in (i)-(v) with the corresponding error maps shown in (vi)-(ix). Here the maps were multiplied by a mask to remove the background and the CSF region. The improvements offered by the proposed scheme can be easily appreciated from the $T_2$ error images and the estimated $T_2$ maps.}
\label{Fig4}
\end{figure*}  

\begin{figure*}[t!]
\centering
\includegraphics[width=1\textwidth]{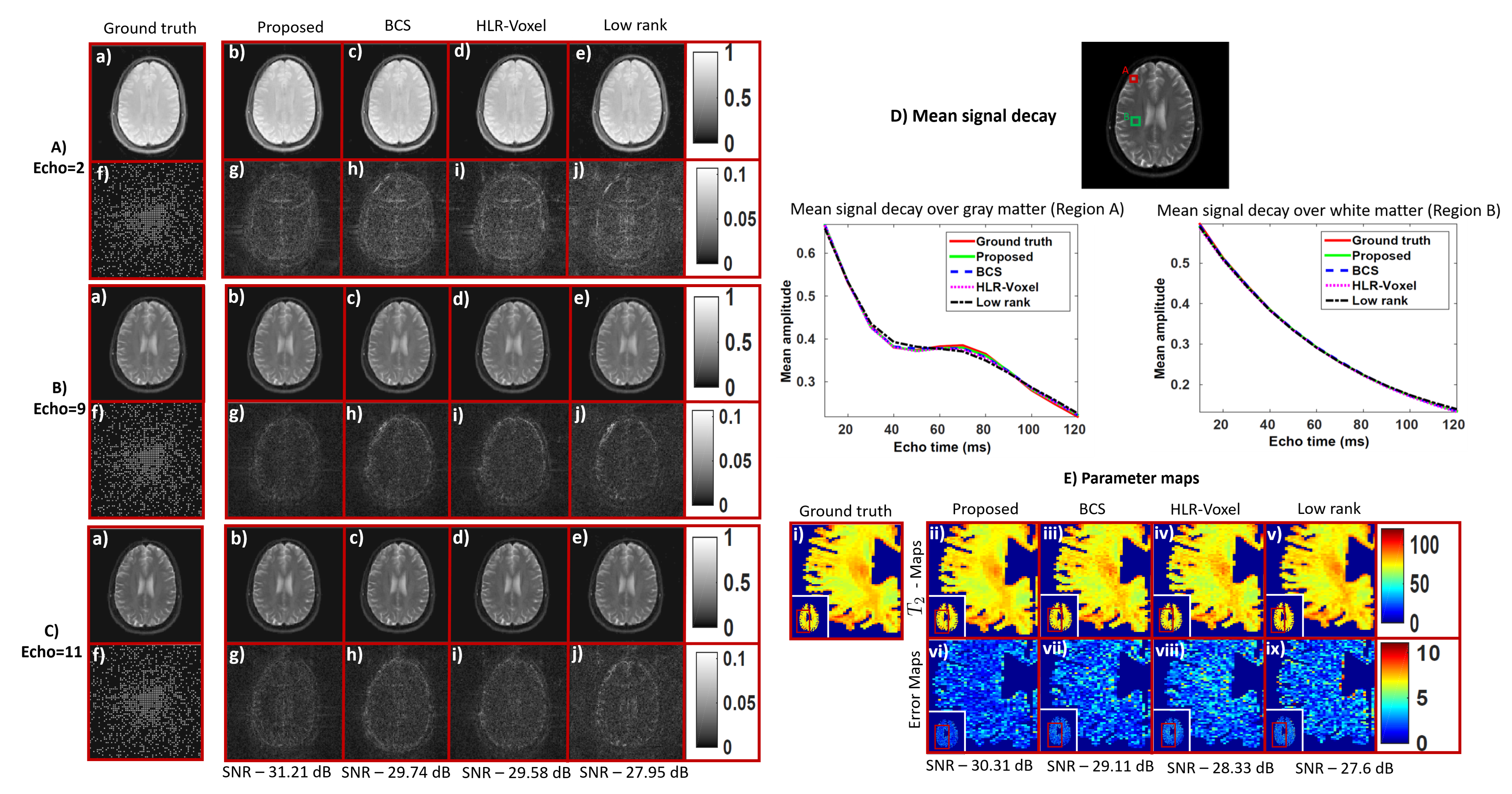}
\caption{Comparison of the proposed method with different reconstruction schemes on the recovery of multi channel data at an acceleration factor of 12: In the first row of A), B) and C) three frames corresponding to 2\rm nd Echo (TE=20ms), 9\rm th Echo (TE=90ms) and 11\rm th Echo (TE=110ms) are shown in (a)-(e) along with the frame of the sampling mask in (f). The corresponding error images are shown in (g)-(j) in the second row of A), B) and C). In D), the mean signal decay is plotted over a gray matter (Region A) and white matter (Region B) region for the ground truth and all the competing methods. The signal within the red ROI was corrupted due to some non-idealities in the acquisition.  In E), the estimated $T_2$ maps, specifically the enclosed rectangular region is zoomed and is shown in (i)-(v) with the corresponding error maps shown in (vi)-(ix). Here the maps were multiplied by a mask to remove the background and the CSF region. We observe that the reconstructions from the proposed method have fewer errors, which can be appreciated from the error maps of the $T_2$ weighted images as well with the noise-like artifacts in the $T_2$ maps.}
\label{Fig5}
\end{figure*}

\begin{figure}[t!]
\centering
\includegraphics[width=0.4\textwidth]{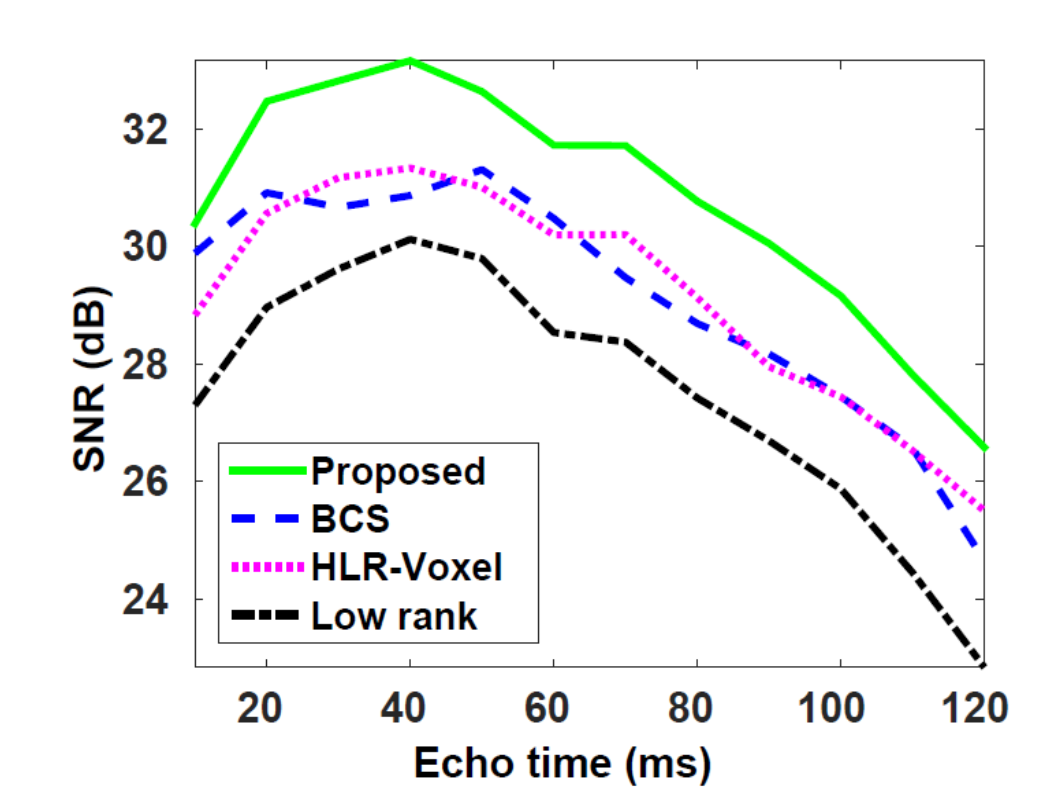}
\caption{Plot showing the SNR of the reconstructions at all the echo times for the proposed and the competing methods. }
\label{Fig6}
\end{figure}

\subsection{Implementation details:}
The details of the alternating minimization algorithm to solve \eqref{eq:x-update} and \eqref{eq:weight-update} are described in Algorithm \eqref{alg:proposed_algo}. We initialize and vary the value of $\epsilon$ as described in \cite{irls}. Specifically, we initialize $\epsilon$ as $\epsilon^{(0)} = \lambda_{max}/100$, where $\lambda_{max}$ is the largest eigen value of the gram matrix $\mathcal{T}(\boldsymbol{\widehat{\rho}}_{init})\mathcal{T}(\boldsymbol{\widehat{\rho}}_{init})^*$, which is formed from the initial guess $\boldsymbol{\widehat{\rho}}_{init} = \boldsymbol{\widehat{\rho}}^{(0)}$. Every iteration, we decrease the value of $\epsilon$ as $\epsilon^{(n)} = \epsilon^{(n-1)}/\gamma$, $\gamma>1$. For all our experiments we set $\gamma=1.4$. We run the optimization algorithm for different values of the regularization parameter $\mu$ and choose the value which results in the best signal to noise ratio (SNR), where $\mbox{SNR}:=20\log\frac{\|\mbf X_g\|_2}{\|\mbf X_{g} - \mathbf{X_{r}}\|_2}$. Here $\mathbf{X_{g}}$ and $\mathbf{X_r}$ are the ground truth and the reconstructed images respectively. To observe the behavior of $\mu$ across different acceleration factors, we reconstructed images at different acceleration factors ranging from six to twelve. We observed that the optimal $\mu$ estimated was fairly constant across different acceleration factors. In the absence of ground truth data it is not possible to compute the SNR. Hence choosing the regularization parameter using the above-mentioned approach is not feasible. In such cases, we could employ heuristic techniques such as the L-curve method \cite{Lcurve} or more sophisticated methods \cite{ramani2008monte, ramani2012regularization}, where risk functions approximating the SNR are used to estimate the optimal regularization parameter. We plan to investigate such techniques for the selection of regularization parameter in the future. We implemented the algorithm in MATLAB, which runs on a Linux workstation with a 3.6Ghz Intel Xeon CPU and 32GB RAM.

\section{Experiments and Results}
We demonstrate our algorithm on a fully sampled axial 2-D dataset, which was acquired on a Siemens 3T Trio scanner with 12 coils using a turbo spin echo sequence. The scan parameters were: TR = 2500 ms, slice thickness = 5 mm, Matrix size = 128$\times$128 and FOV = 22$\times$22 $\mbox{cm}^{2}$. By varying the echo times (TE) from 10 to 120 ms, we acquired $T_2$ weighted images at twelve equispaced TE.
%

\subsection{State-of-the-art methods used for comparison:} We compare the proposed method with three state-of-the-art methods: $k-t$ low rank \cite{ktslr}, blind compressed sensing (BCS) \cite{bhave2015accelerated}, and the pixel-wise structured low-rank prior in \eqref{eq: MORASA-objective}, which we refer to as HLR-Voxel.  An IRLS algorithm \cite{irls} was employed to solve the nuclear norm minimization in \eqref{eq: MORASA-objective}. We compared the methods for both single and multi-channel recovery experiments. 

We also demonstrate the improved speed up offered by the proposed scheme by comparing it with traditional IRLS as well as the multidimensional GIRAF algorithm \cite{GIRAF}. For the traditional IRLS (direct) method, we solve  \eqref{eq:x-update} and \eqref{eq:weight-update} directly without introducing FFT based approximations. After the images are reconstructed, we estimate the $T_2$ maps by fitting a mono exponential model to every pixel.

\subsection{Single channel recovery:} We demonstrate the proposed method on the recovery of single channel $T_2$-weighted data from $30\%$ uniform random measurements and compare it with the three aforementioned methods. To create the single channel data, we performed a principal component analysis (PCA) on the original multi-channel data and selected the most significant component to obtain a coil compressed data. For the proposed method, a filter of size $122\times122\times2$ was used to recover the images. The value of $p=0.6$ for the Schatten norm was chosen for both the $k-t$ low rank and the proposed algorithm. The results from the different methods are shown in Fig. \ref{Fig4}. We observe that the $T_2$ weighted images and the estimated $T_2$ maps from the proposed method have fewer errors and artifacts than the competing methods.

\subsection{Multi-channel recovery:}
In Fig. \ref{Fig5}, we compare the different methods on the recovery of multi-channel $T_2$-weighted data from twelve  fold under-sampled Fourier measurements. The data was retrospectively under-sampled using a combination of uniform Cartesian and a pseudo-random variable density sampling patterns. Specifically, we uniformly undersampled the $x$ and $y$ directions by a factor of 2 and refer this sampling mask as a $2\times2$ uniform Cartesian mask. To increase the incoherence between the frames, we also shifted every frame of the mask by zero or one unit (done randomly) along the $x$ and $y$ directions. We achieved an acceleration factor of twelve by combining the four fold uniform Cartesian mask with a three fold pseudo-random variable density undersampling pattern. Three frames of the sampling mask corresponding to the three echoes are shown in (f)  in Fig. \ref{Fig5} A), B) and C). We used a Schatten $p=0.7$ for both the proposed and the low rank methods. Also  for the proposed method, a filter of size $102\times102\times10$ was used in the recovery of images. Reconstructions corresponding to three echos with TE = 20ms, TE = 90ms and TE = 110ms respectively and the $T_2$ maps for all the methods are shown in A) through C) and E) respectively in Fig. \ref{Fig5}. We observe that the $T_2$ weighted images from the proposed method have fewer errors than the competing methods. Also, the $T_2$ maps corresponding to the proposed method are a lot smoother and have fewer artifacts, especially in the enclosed rectangular region, than those obtained from the competing methods. In Fig. \ref{Fig5} D), we plot the mean signal decay over two regions of interests for all the methods. We observe that the signal corresponding to the proposed method matches closely to the ground truth signal in both cases. We also observe that the mean signal decay from the ROI (in red) in the gray matter is not exactly an exponential function. Nevertheless, the proposed signal model \eqref{eq:signal model} approximates the signal as a linear combination of exponentials and captures the signal decay quite well. This suggests that the performance of the algorithm may degrade gradually, when the signal deviates from an exponential model. In Fig. \eqref{Fig6}, we plot the Signal to Noise ratio (SNR) of the reconstructions at each echo time for the proposed and the competing methods. We observe that for all the methods, the SNR increases for the first few echoes and then decreases for the remaining echoes, which could be due to non-ideal acquisition conditions. Nevertheless, the SNR of the images obtained from the proposed method is higher at all echo times, when compared to the state-of-the-art methods. 

\begin{table}[t!]
  \small
  \centering
  \caption{Effect of filter size on SNR of $T_2$ weighted images.}
  \subfloat[Varying spatial dimension]{%
    \hspace{.2cm}%
    \begin{tabular}{c|r}
        \hline
       filter size       & SNR  \\
        \hline
        128x128x10       & 28.05  \\
        \hline
       122x122x10        &  30.30  \\
        \hline
        114x114x10       & 31.00  \\
        \hline
        108x108x10       & 31.12  \\
        \hline
        \bf{102x102x10}       & \bf{31.21}  \\
        \hline
        100x100x10       & 31.20 \\
        \hline
    \end{tabular}%
    \hspace{.2cm}%
  }\hspace{0,2cm}
  \subfloat[Varying temporal dimensions]{%
    \hspace{.2cm}%
       \begin{tabular}{c|r}
        \hline
         filter size       & SNR (dB)  \\
        \hline
        102x102x11       & 30.80  \\
        \hline
        \bf{102x102x10}       & \bf{31.21}  \\
        \hline
        102x102x7       &  31.13  \\
        \hline
        102x102x4       & 30.96\\
        \hline
        102x102x2       & 30.78  \\
        \hline
        102x102x1       & 29.88  \\
        \hline
    \end{tabular}%
    \hspace{.2cm}%
  }
  \label{tab:table1}
\end{table} 

\subsection{Effect of filter size on image recovery:}

We study the effect of filter size or equivalently the dimensions of the Toeplitz matrix on the SNR of the $T_2$-weighted images recovered from twelve fold under-sampled multi-channel Fourier data in Table \ref{tab:table1}. We study the effect of varying the spatial dimensions of the filter on the SNR in Table \ref{tab:table1}a. We observe that the filters with a smaller spatial support $(102\times102\times10)$ provide improved results than larger filters, thus demonstrating the benefit of exploiting spatial smoothness. A filter with large spatial dimensions $(128\times128\times10)$ fails to exploit any spatial smoothness. We demonstrate the benefit of exploiting the annihilation relations, which takes into account the exponential structure of the signal along the parameter dimension at every pixel, in \ref{tab:table1}b. We observe that a filter having multiple taps along the temporal dimension $(102\times102\times10)$ results in reconstructions with a better SNR than those obtained using a filter with size $(102\times102\times1)$, which just exploits joint sparsity. From these experiments, we also note that varying the spatial support of the filter has a higher impact on the SNR than the temporal support.

\subsection{Effect of approximations on image recovery:}

We study the effect of the approximations, introduced in the proposed method, on the recovery of coil compressed $T_2$-weighted data from $30\%$ uniform random Fourier measurements. We note that the approximations introduced in the proposed method enable efficient computation of the sub-problems \eqref{eq:x-update} and \eqref{eq:weight-update} using Fast Fourier transforms (FFT). This results in a faster convergence to the solution when compared to the IRLS-direct method, as shown in Fig. \ref{Fig2}. Specifically, we observe a 7.5 fold speed up due to the proposed algorithm. In Fig. \ref{Fig3}, we observe that the $T_2$ weighted images and the $T_2$ maps corresponding to the proposed method have similar SNR and image quality compared to those obtained using the IRLS-direct method. These results demonstrate the effectiveness of the approximations introduced in the proposed method. From the figure, we also observe that the images and the maps obtained from the GIRAF algorithm have a lot of errors and artifacts. This is because the approximations in the GIRAF algorithm break down in our setting thus resulting in poor $T_2$ estimates. 

\section{Discussion \& Conclusion}
We introduced a novel structured matrix recovery algorithm to recover an image series with smoothly varying exponential parameters from under-sampled Fourier measurements. As the proposed method exploits the spatial smoothness of the parameters and the exponential structure of the signal along the parameter dimension at every pixel, it results in improved reconstructions over the other state-of-the-art methods. The comparisons on $T_2$ estimation problems in the context of MR parameter mapping demonstrate the potential of this scheme, with reduced errors in both the reconstructed images and $T_2$ maps compared to state of the art methods.

As the size of the filter is not known apriori, we treated it as an optimization parameter and chose the dimension that resulted in the best SNR. We observed that the spatial dimensions of the filter had a greater effect on the SNR than the temporal dimensions. Specifically, a filter with smaller spatial support $(102\times102\times10)$ provided improved reconstructions with higher SNR than a filter of size $128\times128\times10$, which failed to incorporate any spatial smoothness. Hence as the proposed matrix prior incorporates spatial smoothness, it eliminates the need for additional spatial regularizers or priors to further constrain the image recovery. Similarly, the reconstructions using a filter with multiple taps along the temporal dimension $(102\times102\times10)$ had higher SNR than the filter with one tap, which exploits the joint sparsity of the Casorati matrix formed from the Fourier samples. We also observed that for both the multi-channel and coil combined data, the filter sizes yielding reconstructions with highest SNR were different. The size of the filter reflects the complexity of the model which is usually dependent on the number of measured samples, number of coils etc. Since the model complexity is different for both the datasets, it resulted in different sizes of the filter.

To solve the optimization problem, we employ an iterative least squares (IRLS) based strategy, which decouples the original problem into two sub-problems. We adopt a hybrid approach to keep the memory demand and computational complexity low. Specifically, we introduce novel approximations, which allow us to solve the sub-problems using FFTs. This resulted in a faster convergence to the same solution as the one obtained using the IRLS (direct) method. The proposed algorithm was approximately 7.5 times faster than the IRLS (direct) method. We also observed that the GIRAF algorithm broke down in our setting resulting in poor reconstructed images and $T_2$ maps.

The proposed framework may be extended to the multi-dimensional parameter setting. For instance, in MR parameter mapping, if the signal along the parameter dimension at every pixel varies as a function of both $TR$ and $TE$, then the filter coefficients will be dependent on both exponential parameters $T_1$ and $T_2$. Hence, annihilation relations similar to \eqref{eq:annihilation 1D} and \eqref{eq:annihilation 3D} can be derived, which can be compactly represented using a low rank Toeplitz matrix. The low rank property of the Toeplitz matrix can then be enforced to recover the images from under-sampled Fourier measurements. We plan to investigate this problem in the future.

The results in Fig.\eqref{Fig5}.D indicate that the performance of the algorithm degrades gradually, when the voxel time profiles deviate from the exponential signal model. Specfically,  such signals may be reasonably approximated as a linear combination of few exponentials. In addition, the use of the low-rank penalty that only requires the matrix to be approximately low-rank, rather than a low-rank constraint, also allows the signal  to deviate from the exponential model. We did not account for the presence of artifacts due to stimulated echoes  \cite{ben2015rapid, majumdar1986errors1, majumdar1986errors2, crawley1987errors, hennig1988multiecho} which are formed when a turbo spin echo sequence is used. Further investigation is required to study the impact of such errors.


\bibliographystyle{IEEEtran}
\bibliography{ref}

\end{document}